\documentclass[letterpaper, 10 pt, conference]{ieeeconf}  

\usepackage{graphicx}
\usepackage{svg}
\usepackage{hyperref}
\usepackage{amsmath}
\usepackage{cite}
\usepackage{color}

\def\baselinestretch{0.984}

\usepackage{booktabs}
\usepackage{etoolbox}
\makeatletter
\patchcmd{\@makecaption}
  {\scshape}
  {}
  {}
  {}
\makeatletter
\patchcmd{\@makecaption}
  {\\}
  {.\ }
  {}
  {}
\makeatother

\IEEEoverridecommandlockouts                              

\usepackage{amsmath} 

\DeclareMathOperator*{\argmin}{argmin}

\title{\LARGE \bf
SafetyNet: Safe planning for real-world self-driving vehicles\\using machine-learned policies}

\author{Matt Vitelli$^*$, Yan Chang$^*$, Yawei Ye$^*$, Maciej Wołczyk, Błażej Osiński, \\ Moritz Niendorf, Hugo Grimmett, Qiangui Huang, Ashesh Jain, Peter Ondruska$^+$\\
\thanks{$^*$ Equal contribution.}
\thanks{$^+$ Authors are with Toyota, Woven Planet, Level 5 self-driving division \{matthew.vitelli, yan.chang, yawei.ye, maciej.wolczyk, blazej.osinski, moritz.niendorf, hugo.grimmett, qiangui.huang, ashesh.jain, peter.ondruska \}@woven-planet.global. }
\thanks{Videos are available at \url{safety.l5kit.org}}
}

\begin{document}

\maketitle
\thispagestyle{empty}
\pagestyle{empty}


\begin{abstract}
In this paper we present the first safe system for full control of self-driving vehicles trained from human demonstrations and deployed in challenging, real-world, urban environments. Current industry-standard solutions use rule-based systems for planning. Although they perform reasonably well in common scenarios, the engineering complexity renders this approach incompatible with human-level performance. On the other hand, the performance of machine-learned (ML) planning solutions can be improved by simply adding more exemplar data. However, ML methods cannot offer safety guarantees and sometimes behave unpredictably. To combat this, our approach uses a simple yet effective rule-based fallback layer that performs sanity checks on an ML planner's decisions (e.g. avoiding collision, assuring physical feasibility). This allows us to leverage ML to handle complex situations while still assuring the safety, reducing ML planner-only collisions by 95\%. We train our ML planner on 300 hours of expert driving demonstrations using imitation learning and deploy it along with the fallback layer in downtown San Francisco, where it takes complete control of a real vehicle and navigates a wide variety of challenging urban driving scenarios.
\end{abstract}
%

\section{Introduction}

Self-Driving Vehicles (SDVs) have the promise to revolutionize several industries including people and goods transportation. However, the development of L4+ SDVs has proved to be a significant challenge. Today, the main bottleneck is the vehicle's ability to safely handle the `long tail' of driving events \cite{jain2021autonomy}. World-class SDVs can handle common situations, but can behave unsafely in the many, rarely-occurring scenarios that are encountered on the road. 

In the self-driving stack, the \emph{planning} module is most responsible for this bottleneck. It determines what the SDV should do in any given situation. A traditional \emph{rule-based} planning approach selects a trajectory that minimizes a hand-engineered loss function. In order to improve its performance, engineers must design new terms in that loss function or re-tune their respective weights, for each driving scenario. This process is expensive and scales poorly to new geographies. Unlike perception, planning has benefited little from modern machine learning techniques, which leverage large quantities of data in order to avoid the hand-engineering of rules.

\begin{figure}
    \centering
    \includegraphics[width=\linewidth]{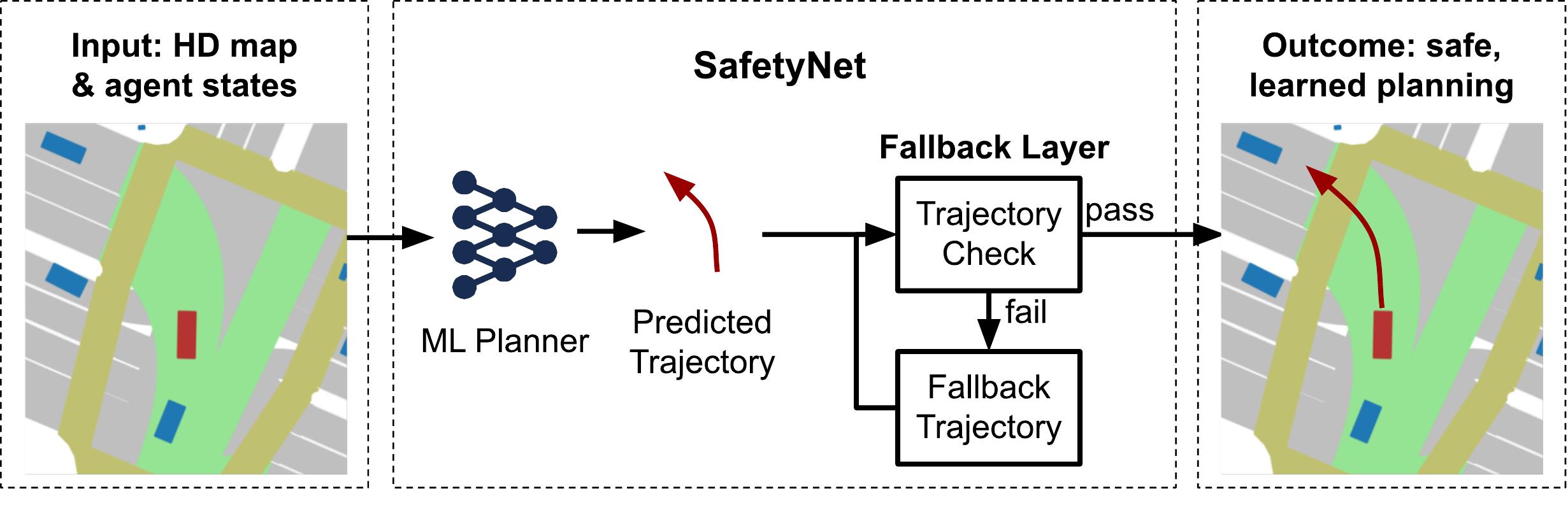}
    \includegraphics[width=\linewidth]{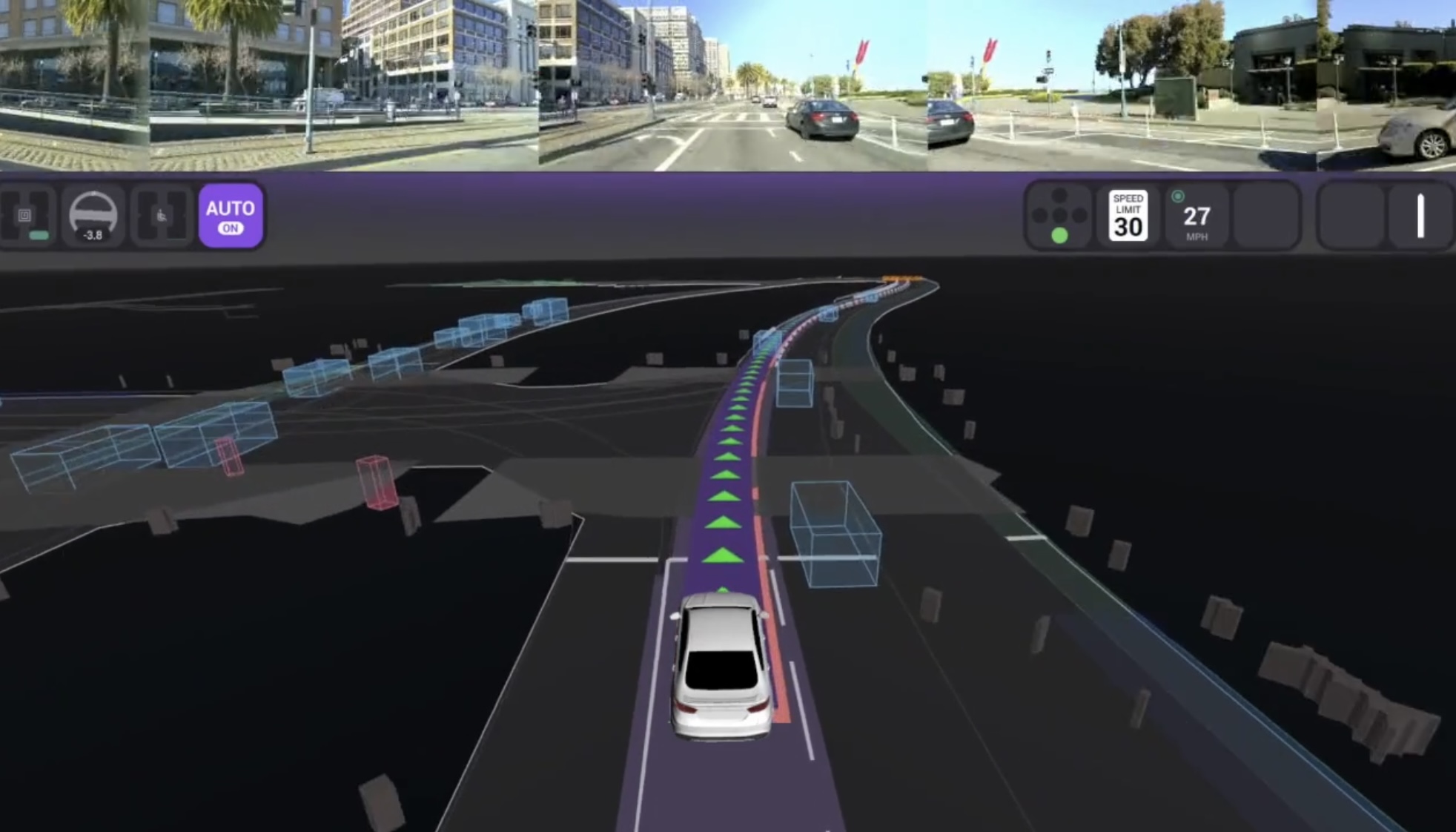}
    \caption{\emph{Top:} SafetyNet is the best-of-both combination of ML planning that improves with data, and rule-based safety and legality guarantees from the fallback layer. \emph{Bottom: } an example of SafetyNet deployed to control a real-world self-driving vehicle on the streets of San Francisco.} 
    \label{fig:system}
    \label{fig:intro}
\end{figure}

Recently, the work of \cite{bansal2018planning-5} has demonstrated the first machine-learned policies for autonomous driving learned directly from human demonstrations. These approaches, although they scale much better than the hand-engineering method, do not provide the interpretability and safety guarantees required to safely deploy these systems in production.

In this work we propose SafetyNet: the first autonomous driving system to combine the strengths of an ML planner with the interpretable safety of a rule-based system, road-tested in busy San Francisco. The ML component is a high-capacity planning policy trained from expert demonstrations, and its performance scales with the amount of training data without the need for costly behavior engineering. To improve system safety, decisions of the ML planner pass through a lightweight \emph{fallback layer}: a simple, rule-based system that tests the decisions against a small set of checks, and can minimally modify them to improve safety if required. This allows SafetyNet to transparently enforce safety and legality constraints, such as avoiding collisions, road rules violations, while simultaneously maximizing comfort metrics.

This combination outperforms ML-only systems and allows us to \emph{safely} deploy an ML planning system in the busy streets of San Francisco, constituting the first demonstration of its kind. Our system exhibits a variety of maneuvers such as lane-following, keeping the distance to other vehicles, and navigating intersections without impeding the safety of the vehicle and other traffic participants.

Our contributions are three-fold:
\begin{enumerate}
    \item The first combination of machine learning and a lightweight hand-engineered system to control a self-driving vehicle that learns from data while offering safety and legality guarantees.
    \item The first evaluation of such a system in the challenging, real-world, urban environment of downtown San Francisco.
    \item The source code will be provided to the public to encourage the advancement of the field\footnote{The source code for the ML planner and a reference implementation of the fallback layer will be available at \url{safety.l5kit.org}.}.
\end{enumerate}

\section{Related works}

\begin{figure}
    \centering
    \includegraphics[width=\linewidth]{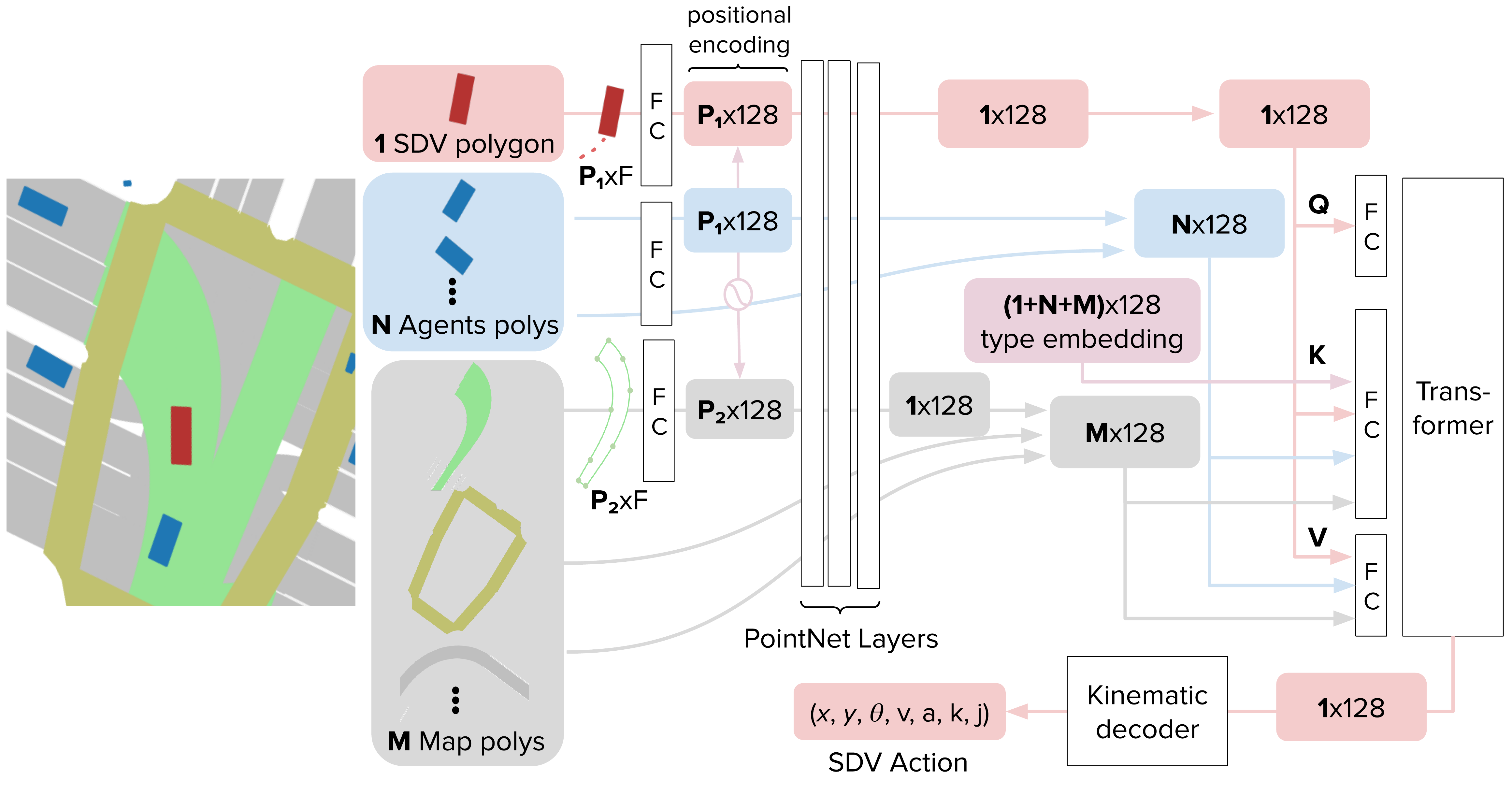}
    \caption{The neural network architecture of the presented ML planning model is inspired by VectorNet \cite{Gao_2020_CVPR}. The vectorized information on each agent and map element is encoded by a PointNet network. This local information is combined by a Transformer into global embedding. The embedding is later translated into actions via kinematic decoder.}
    \label{fig:model}
\end{figure}

\textbf{Trajectory optimization-based planning.} Traditional trajectory optimization-based planning systems are widely used in both academia and industry\cite{montemerlo2009, Buehler2009, paden2016survey, fan2018}.
Here the motion planning task is formulated as an optimization problem, usually by hand-engineering a cost function. The optimal trajectory is then generated by minimizing this cost using optimization algorithms, such as A* search-based methods \cite{Buehler2009, Ajanovic2018}, sampling-based methods \cite{KuffnerRRT-connect2000, karaman2011,Bandyopadhyay13}, dynamic programming \cite{fan2018} or combinations thereof that decompose the problem in a hierarchical fashion \cite{wei2014behavioral}.

However, it is very difficult to hand-craft an objective function that provides a human-like trade-off between comfort, safety, and route progress over a wide variety of driving situations\cite{chang2021systems}. In comparison, encoding certain hard constraints, e.g. obstacles avoidance, physical feasibility, requires far less engineering.
Therefore, building a lightweight hand-engineered system whose only purpose is to detect and correct infeasible trajectories is much simpler and more scalable.

Additionally, these hand-engineered approaches do not improve with data, and their performance does not generalize well in highly unstructured urban scenarios. Therefore, tremendous engineering efforts are needed to fine-tune them, in particular when expanding to a new operational design domain (ODD) or new geographies.

\textbf{Machine-learned planning.} Recently, ML planning has gained attention due to successes in deep learning. This approach has the advantage of avoiding hand-crafted rules and scales well with data, thus performing better and better as more data is used for training. Therefore, this approach has great potential to handle a wide variety of driving situations \cite{bansal2018planning-5, bojarski2016end2end, zeng2019end, Hawke2020UrbanDW}. Next, we introduce the two most commonly used ML paradigms for motion planning.

\textbf{(1) Imitation learning (IL).} IL is a supervised learning approach in which a model is trained to mimic expert behavior. The first application of IL to autonomous driving was the seminal ALVINN\cite{Pomerleau1989} back in 1989, which mapped the sensor data to steering and performed rural road following.
More recently, \cite{bojarski2016end2end, Hawke2020UrbanDW} demonstrated end-to-end driving using multiple-camera input alone, but the real-world driving results are limited to simple tasks such as lane follow or urban driving with light traffic.
ChauffeurNet\cite{bansal2018planning-5} proposed to apply IL on a bird's eye view of a scene and use synthetic perturbations to alleviate the covariate shift problem \cite{Ross2011ARO}, but it is yet to be tested in real-world urban environments.

\textbf{(2) RL \& IRL.} Reinforcement learning (RL) is well-suited for sequential decision processes such as self-driving as it handles the interaction between the agent and the environment.
Several methods\cite{shalev2016safe, riedmiller2007learning, kendall2019learning} have been proposed to apply RL to autonomous driving.
In particular \cite{shalev2016safe} proposes combining learned and rule-based components, similarly to us, but the reported results are only from simulation.
On the other hand, inverse reinforcement learning (IRL)\cite{ziebart2008maximum, Wulfmeier2015MaximumED} is another popular ML paradigm applied to autonomous driving, which infers the underlying reward function based on expert demonstrations as well as a model of the environment.
However, all these methods are yet to be evaluated in real-world urban driving.

The ML planning approaches introduced above, although very promising, do not provide safety guarantees, which prevent them to be deployed at scale in the real world. We are inspired by this paradigm but aim to mitigate this limitation by the SafetyNet proposed in this paper. 

\textbf{Hybrid approaches.}
The combination of ML and traditional motion planning techniques falls mostly into two categories:
\emph{ML-based heuristics}, which are leveraged to improve traditional planning algorithms, e.g. in terms of speed up  \cite{qureshi2019MPNetworks,arslan2015GuidingRRT,chaing2019RLRRT,pulver2020pilot}.
\emph{Modular approaches}, where expert planners are leveraged to generate the trajectory candidates, e.g., by evaluating trajectories against a ML-based cost volume \cite{zeng2019end, casas2021mp3}.
The latter of these works, \cite{casas2021mp3} also provides safety guarantees based on imposing a very high cost on trajectories leading to a potential collision. These safety guarantees were not however verified in the real world.

Another specific area of research that has emerged in this field is the study of safety frameworks \cite{Shalev2017RSS, althoff2014Reachability}.
While this work is relevant, our goal is not to propose a comprehensive framework for safety, but rather a simple yet effective method that allows for the deployment of a powerful neural network planner that learns and improves with data while ensuring certain safety and legality constraints. 

SafetyNet leverages the strengths of the expert system to guarantee certain determinism, legality, and safety rules for specific scenarios while relying on the machine-learned motion planner for nominal trajectory generation. 

\section{Hybrid ML planning system}

In this section we describe SafetyNet, our system for combining a machine-learned motion planner with an effective fallback layer to deliver a trajectory planning system for SDVs. Instead of relying on hand-engineered driving rules, this system learns to drive from expert driving demonstrations while guaranteeing certain interpretable safety constraints, such as avoidance of collisions and adherence to traffic rules.

The SafetyNet system is outlined in Fig.~\ref{fig:system}. It is composed of an ML neural policy network (``ML Planner'') $\mathcal{M}$ that takes as input $I$ the environment state around the SDV and produces an intended trajectory $\bar{\tau}$ to be taken by SDV. This trajectory is validated to obey the required constraints and in the case it does not satisfy them the closest trajectory $\tau^i$ is taken from a set of safe trajectory candidates.

\subsection{Input and Output}
\textbf{Input representation.}
The input data is encoded in an ego-centric frame of reference where the SDV is always at a fixed location relative to a frame.
As shown in Fig.~\ref{fig:model}, the input to our model consists of: 
\begin{enumerate}
\item SDV: the current and past poses of the SDV and its size.
\item Agents: the current and past poses of perceived agents, their sizes, and object type (e.g. vehicle, pedestrian, cyclist) produced by the SDV's perception system.
\item Static map elements: road network from High Definition (HD) maps including lanes, cross-walks, stop lines, localized using the SDV's localization system.
\item Dynamic map elements: traffic light states, and static obstacles detected by the perception system (e.g. construction zones).
\item Route: the intended global route that the SDV should follow. 
\end{enumerate}
We use a vectorized input representation based on~\cite{Gao_2020_CVPR} to encode the perception outputs and map elements to vector sets.
Each element includes a pose relative to the SDV pose, as well as additional features, such as the element type, time of observation.

\textbf{Output representation.}
\label{section:output representation}
We define a trajectory {$\tau $} as a sequence of {$T$} discrete states separated uniformly in time by {$\Delta t$}. Each state {$s_t$} is defined as:
\begin{equation}
\label{state_space}
s_t = \{x_t, y_t, \theta{_t}, v_t, a_t, k_t, j_t\}.   
\end{equation}

where {$x_t, y_t, \theta{_t}$} correspond to the pose of the rear axle of the SDV w.r.t. a fixed coordinate frame at time {$t$} and {$v_t, a_t, k_t, j_t$} correspond to the velocity, longitudinal acceleration, curvature and jerk respectively.

\subsection{ML planner}

\begin{figure}[t]
    \centering
    \includegraphics[width=\linewidth]{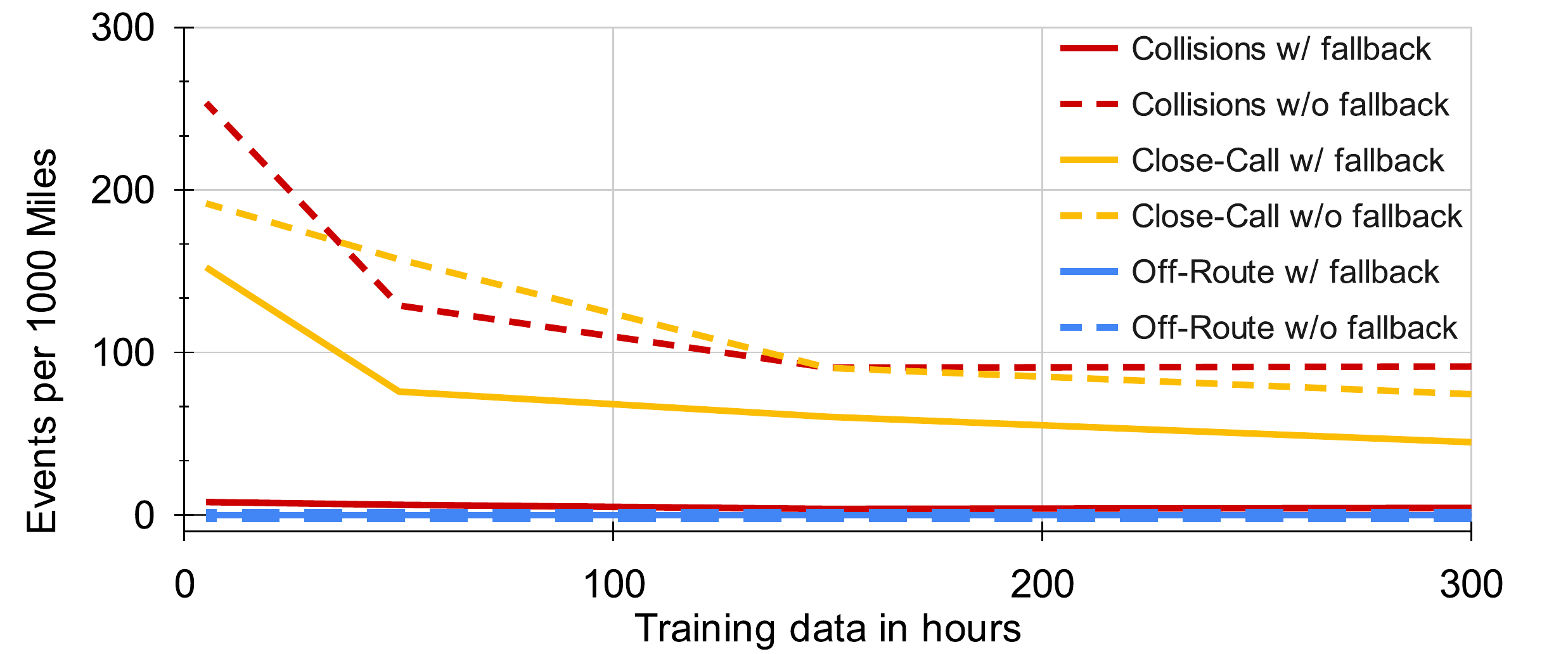}
    \includegraphics[width=\linewidth]{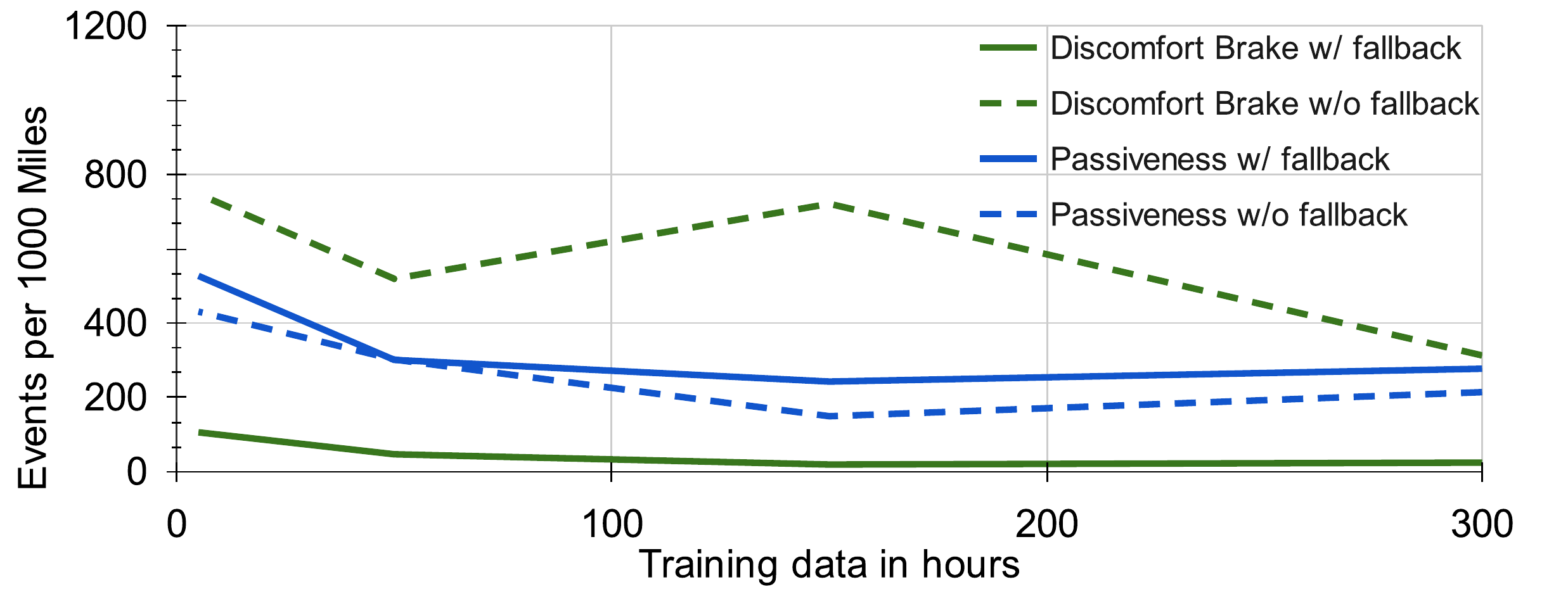}
    \caption{Safety (\emph{top}) and comfort (\emph{bottom}) events as a function of dataset size for the ML planner (with and without the fallback layer). We see that the addition of the fallback layer significantly improves every metric, save passiveness. We note that the number of collisions is not decreasing quickly as the ML planner is trained on more data, but despite this the fallback layer ensures a high level of safety.}
    \label{fig:collisions_closed_loop}
\end{figure}

\begin{table}[t]
  \centering
  \def\arraystretch{1.1}
  \begin{tabular}{@{}c c c c c@{}}
        \hline\toprule
        & \multicolumn{4}{c}{\textbf{Average Displacement Error [m]}} \\
        \textbf{Training set size} & \textbf{@1s} & \textbf{@2s} & \textbf{@3s} & \textbf{@4s}\\
        5h & 0.24 & 0.90 & 1.70 & 2.52\\
        50h & 0.14 & 0.51 & 0.98 & 1.53\\
        150h & 0.12 & 0.40 & 0.77 & 1.22\\
        300h & 0.11 & 0.37 & 0.73 & 1.18\\
     \bottomrule\hline
    \end{tabular}
    \caption{\label{tab:open loop} Open-loop performance trend with increasing training set size. The open-loop prediction error improves when trained with more data.}
\end{table}

The ML planning component of our system takes the input $I$ capturing the states around the SDV and outputs the trajectory $\bar{\tau}$ to be executed.

\textbf{Model architecture.}
Inspired by \cite{Gao_2020_CVPR}, our model is built on a hierarchical graph network-based architecture as shown in Fig.~\ref{fig:model}.
It consists of a PointNet-based~\cite{qi2016pointnet} local subgraph for processing local information from vectorized inputs and a global graph using a Transformer encoder~\cite{NIPS2017_3f5ee243} for reasoning about interactions over agents and map features.

To ensure the predicted trajectories are physically feasible, we introduce a kinematic decoder, which models the vehicle kinematic using a unicycle model~\cite{paden2016survey}. First, a 3-layer multilayer perceptron (MLP) is added after the Transformer encoder, which predicts longitudinal jerk {$j_{1:T}$} and curvature {$k_{1:T}$} for each time step within the prediction horizon {$T$}. Then a kinematic layer takes the predictions as well as the current ego state to roll out the next state of the ego:
\begin{equation} \label{eqn:1}
s_{t + 1} = f(s_t, k_t, j_t, \gamma),
\end{equation}
where ${f}$ is the update function of the kinematic model, ${\gamma}$ comprises a set of parameters for vehicle kinematic constraint, including the maximum allowed jerk, acceleration, curvature and steering angle, which are used to clip controls to ensure physical feasibility.

Inspired by the deep kinematic model introduced in ~\cite{cui2020deep}, we implement ${f}$ as follows:
\begin{equation}
s_{t + 1} = s_{t} + \dot{s_{t}}\Delta{t},
\end{equation}

where the state derivatives are computed as follows:
\begin{equation}
\begin{split}
&\dot{x_{t}} = v_{t} \cos{\theta{_t}}, \; \dot{y_{t}} = v_{t} \sin{\theta{_t}}, \\
&\dot{\theta{_t}} = {k_t} v_{t}, \;
\dot{v_t} = a_{t},\; \dot{a_t} = j_{t}. \\
\end{split}
\end{equation}

\textbf{Training framework.}
We use imitation learning to train a driving policy that mimics expert driving behavior by minimizing the L1 loss between the poses generated by the model and the ground truth poses. Following \cite{bansal2018planning-5}, we include perturbations to extend the distribution of states seen during the training and thus reduce the impact of the covariate shift \cite{Pomerleau1989, Ross2011ARO}. Although previous work used a pre-solver to smooth the target trajectory after applying perturbations, we can skip that thanks to the fact that we are using a kinematic decoder. Instead, we can simply penalize large values of jerk and curvature to reduce jerk and improve driving comfort.
The final loss is then:
\begin{equation}
    L = \sum_{t=1}^T \| p_t - \hat{p}_t \|_1  + \alpha \| k_t \|_2 + \beta \| j_t \|_2,
\end{equation}
where $p_t$ is the predicted pose $(x_t, y_t, \theta_t)$ at time $t$, $\hat{p_t}$ is the target pose, and $\alpha$ and $\beta$ are hyperparameters.

\subsection{Fallback Layer}
After generating an ML trajectory, our system evaluates it along several dimensions for dynamic feasibility, legality, and collision probability, and determines a trajectory label \{\textit{Feasible}, \textit{Infeasible}\}. We describe these next.

\textbf{Dynamic feasibility.}
We evaluate whether the input trajectory remains within a feasible envelope characterized by the SDV's dynamics limits. Concretely, we evaluate each trajectory state and check whether the parameters, including longitudinal jerk, longitudinal acceleration, curvature, curvature rate, lateral acceleration, and steering jerk (curvature rate $\times$ velocity) are within reasonable bounds.

The bounds for those parameters were obtained from real-world vehicle testing. In practice, we typically use more conservative limits for jerk, longitudinal acceleration, and lateral acceleration to remain within comfortable limits.

\textbf{Legality.}
For a given trajectory, we evaluate whether it is violating the traffic rules. A trajectory will be labeled as \textit{Infeasible} if any of the following violations happens:
\begin{itemize}
\item Running the stop sign,
\item Violation of the right of way,
\item Running a red traffic light,
\item Leaving the drivable surface.
\end{itemize}

\textbf{Collision likelihood.}
We check each state in the given trajectory for collisions with predicted poses of other agents from an in-house prediction module.
Collision detection is performed by rasterizing future agent predictions and checking for overlaps with planned ego poses.
Additionally, we also check for longitudinal distance, time-to-collision, and time headway violations along the trajectory. If any of the collision likelihood checks fail, the trajectory is labeled as \textit{Infeasible}.

\textbf{Fallback trajectory generation.}
Assuming the ML trajectory is labeled as \textit{Feasible}, we will directly execute it. If the trajectory is labeled as \textit{Infeasible}, we select a feasible fallback trajectory as close as possible to the ML trajectory.

For this we use a trajectory generation method based on~\cite{werling2012optimal}, generating a number of lane-aligned trajectory candidates $\tau^i$. These candidates consist of speed keeping, distance keeping, and emergency stopping maneuvers.
Our implementation can be easily adapted to specific scenarios of interest.

Each of the generated trajectories is checked for feasibility as described above and the trajectory candidate which is most similar to the ML trajectory is selected for execution:
\begin{equation}
    \tau = \argmin_{\tau^i} \| \bar{\tau} - \tau^i \|_2.
\end{equation}

\section{Experiments}

\begin{table*}[t]
   \def\arraystretch{1.3}
   \centering
    \begin{tabular}{ @{} l c c c c c @{}  }
        \hline\toprule
         & \multicolumn{5}{c}{\textbf{\# events per 1k miles}} \\

         \textbf{Planner Type} & \textbf{Collisions} & \textbf{Close Calls} & \textbf{Discomfort Braking} & \textbf{Passiveness} & \textbf{Deviation from Route}\\

        ML Planner & 91.5 & 74.5 & 312.8 & 213.9 & 0.0 \\
        
         ML Planner + Fallback Layer & 4.6     & 45.0 & 24.2 & 277.0 & 0.0 \\

        \emph{Change Rate}  & -95\% & -40\% & -92.3\% & +29.5\% & 0\% \\
         \bottomrule\hline
         \vspace{-3mm}
    \end{tabular}
\caption{\label{tab:fallback layer}Comparing the performance of an ML-Planner-only approach vs SafetyNet, in closed-loop evaluation. This shows that SafetyNet significantly reduces collisions, close calls, and discomfort breaking, at the expense of more passiveness. See Fig.~\ref{fig:fallback_cases} for some qualitative examples.}
\vspace{-5mm}
\end{table*}

In this section we evaluate our system across several dimensions: (a) the performance of the ML planner in simulation when trained on an increasing amount of data (no fallback layer), (b) the effectiveness of the fallback layer in simulation, and (c) the performance of SafetyNet in the real world when controlling a real vehicle in San Francisco. We start by describing the datasets and metrics used during the evaluation.

\subsection{Data}
We created an in-house dataset to train and evaluate our system: 380 hours of urban driving collected in Palo Alto and San Francisco. It contains a wide range of driving scenarios in a densely populated urban environment. The dataset consists of 25 second \emph{scenes}, which capture the perception output, the SDV trajectory, and HD maps. We partition the dataset into 300h for training and 80h for testing.

\subsection{Metrics}
We validate our planner using large-scale real-world driving dataset with closed-loop simulation. During simulation, we execute the planner and motion controller modules and simulate the SDV motion. The vehicle model is calibrated to the real-world SDV. Since the simulated SDV pose can diverge significantly from the logged pose in the dataset, we allow the other road \emph{agents} to be reactive in their longitudinal behavior, avoiding collisions while preserving their trajectories from the dataset.

To evaluate the closed-loop performance of the model, for each scene in the test set we simulate the SDV by unrolling the driving policy for the full duration of the scene and detect the following binary events:
\begin{enumerate}
    \item \emph{Collisions}: the simulated SDV is $<$5cm from the road boundaries, static obstacles, or agents.
    \item \emph{Close-calls}: the simulated SDV has no collision, but either gets within 25cm of another agent, has a time-to-collision $<$1.5s, or has a time headway to another agent $<$1s.
    \item \emph{Discomfort braking}: the simulated SDV’s jerk drops below $-5$m/s$^3$.
    \item \emph{Passiveness}: the simulated SDV travels slower than its behavior in the dataset by $<-$5m/s and it is spatially behind its dataset position.
    \item \emph{Off road events}: the simulated SDV deviates from the dataset route center line by $>$10m.
\end{enumerate}
After identifying the events in each scene, we aggregate them, normalizing by the number of miles driven in the simulations. All metrics are reported as the total number of events per 1000 miles.

Additionally, to evaluate the open-loop performance, we compute the Average Displacement Error (ADE) between the position of the simulated SDV and the dataset position. This gives us insight into trajectory similarity between simulation and the dataset (see Table~\ref{tab:open loop}).

Finally, we test how the system performs in the real world by deploying it in downtown San Francisco. We evaluate the ML planner performance via how often the fallback trajectory is used and we provide qualitative examples. 

\subsection{Effect of dataset size}
For this experiment, the fallback layer is disabled, and the ML planner's trajectory is always executed. We evaluate the performance of various ML planners trained with varying dataset sizes: \{5, 50, 150, 300\} hours in simulation.

Looking at the dashed lines in Fig.~\ref{fig:collisions_closed_loop} (w/o fallback) we observe that the closed-loop performance across all safety and comfort metrics increases with training set size, and that more data should continue improving performance, albeit slowly. Models trained on $<$50h produce unstable driving policies that have significantly more collisions, off-route events, or even cases where the SDV remains completely still (passiveness). When dataset size is increased to 300 hours, performance improves significantly and the learned driving policy is able to reduce collisions, close-call and discomfort brakes, and reliably follows the route. In terms of passiveness, performance plateaus, which we attribute to causal confusion inherent to open-loop training \cite{dehaan2019causal}. Similarly, the average displacement error (ADE) of the open-loop predictions keeps improving when trained with more data, as shown in Table \ref{tab:open loop}.

\subsection{Effect of fallback layer}
\label{sec:performance}

\begin{figure}
    \centering
    \includegraphics[width=\linewidth]{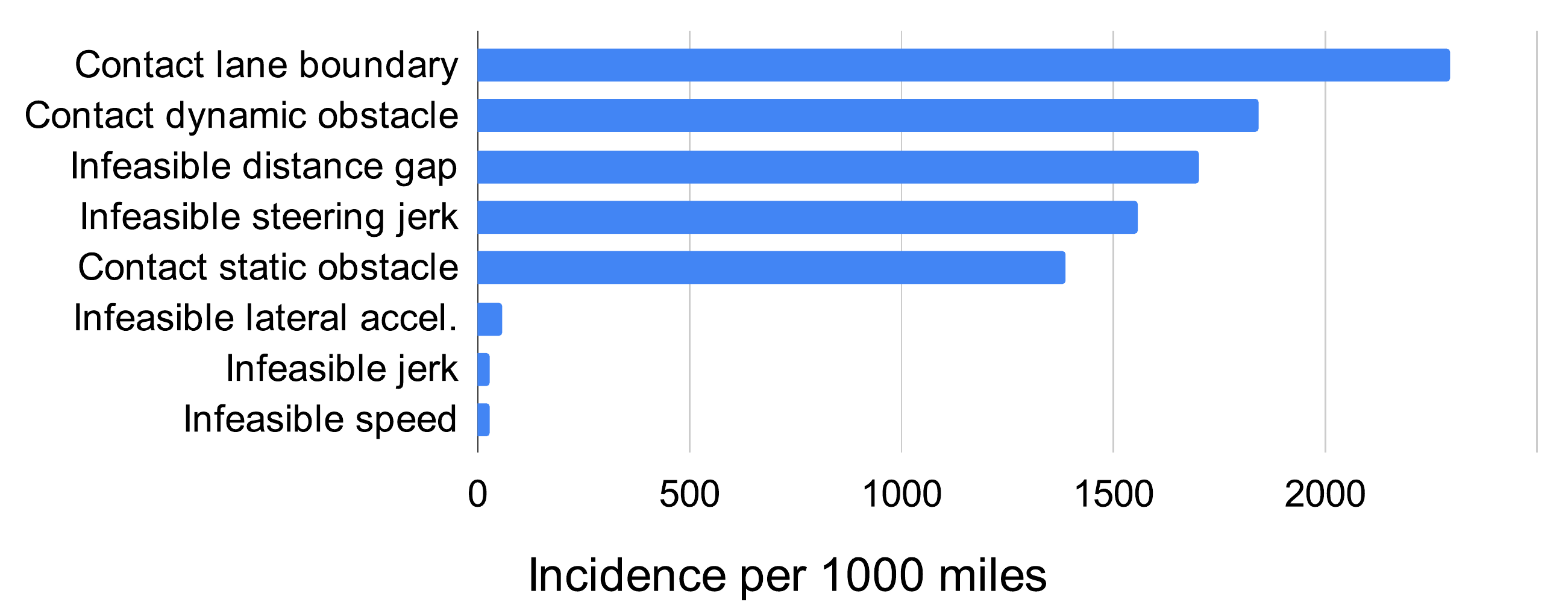}
    \caption{The distribution of causes for a fallback trajectory to be used over the ML Planner trajectory.}
    \label{fig:fallback_trigger}
    \vspace{-3mm}
\end{figure}

\begin{figure}
    \centering
    \includegraphics[width=\linewidth]{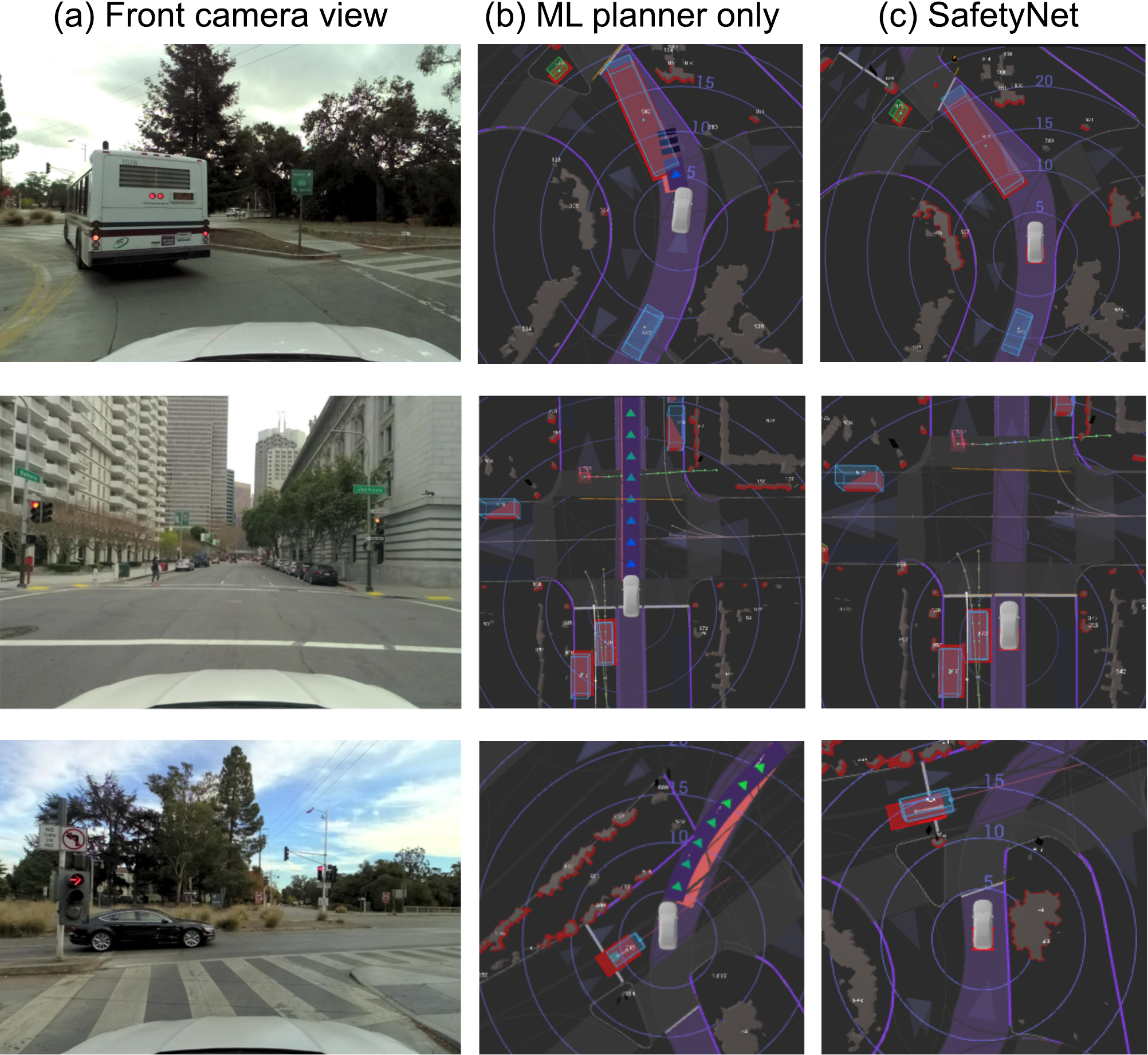}
    \caption{Examples of when the fallback layer prevented unsafe behavior caused by the ML planner. \emph{[Top]} avoiding collision with a bus, \emph{[middle]} running a red light, and \emph{[bottom]} ensuring yielding to vehicle with right-of-way. We report the unsafe trajectory in column (b), and SafetyNet safe trajectory in column (c).}
    \label{fig:fallback_cases}
    \vspace{-4mm}
\end{figure}

Now we evaluate the effectiveness of the fallback layer in simulation by comparing the SafetyNet performance with, and without the fallback layer enabled. Here the ML planner is trained on the full 300h dataset. As shown in Tab.~\ref{tab:fallback layer}, collisions are reduced by 95\%, close call events by 40\%, and discomfort braking by 92\% when the fallback layer is enabled. This clearly demonstrates the value of the fallback layer and its importance for real world deployment. Crucially, we see that the solid lines in Fig.~\ref{fig:collisions_closed_loop} (w/ fallback) are close to zero regardless of the maturity of the ML planner. This indicates that, with our method, we can safely deploy not only well-performing ML planners, but even immature ones (trained on $<$50h of data), thus facilitating faster development and evaluation cycle. 

We note that there is also a 29.5\% increase in passive behavior when the fallback layer is enabled, due to the fact that the SDV drives more conservatively. We see this as a necessary trade-off for reducing collisions and close calls.

In Fig.~\ref{fig:fallback_trigger} we show the distribution of events that trigger the fallback trajectory to be used during simulation. The main causes for fallback behaviors are the ML trajectory leaving the drivable surface, contacting dynamic agent predictions, infeasible distance gap with other agents, infeasible steering jerk, and contact with static obstacles. By manually triaging these ML planner failure cases, we see (a) contact lane boundary issues are a result of the ML planner cutting corners too closely, and (b) collisions are caused by the network not paying enough attention to the size of the large vehicles and assuming that vehicles are of similar size (Fig.~\ref{fig:fallback_cases}, \emph{top}). Moreover, near traffic light intersections the ML planner occasionally generates trajectories that do not properly yield to oncoming traffic (Fig.~\ref{fig:fallback_cases}, \emph{bottom}). It is caused by the synthetic perturbations applied to the ego poses during training. Note that in all these examples, the fallback layer successfully prevented collisions from happening.

\subsection{Failure cases}
\begin{figure}
    \centering
    \includegraphics[width=\linewidth]{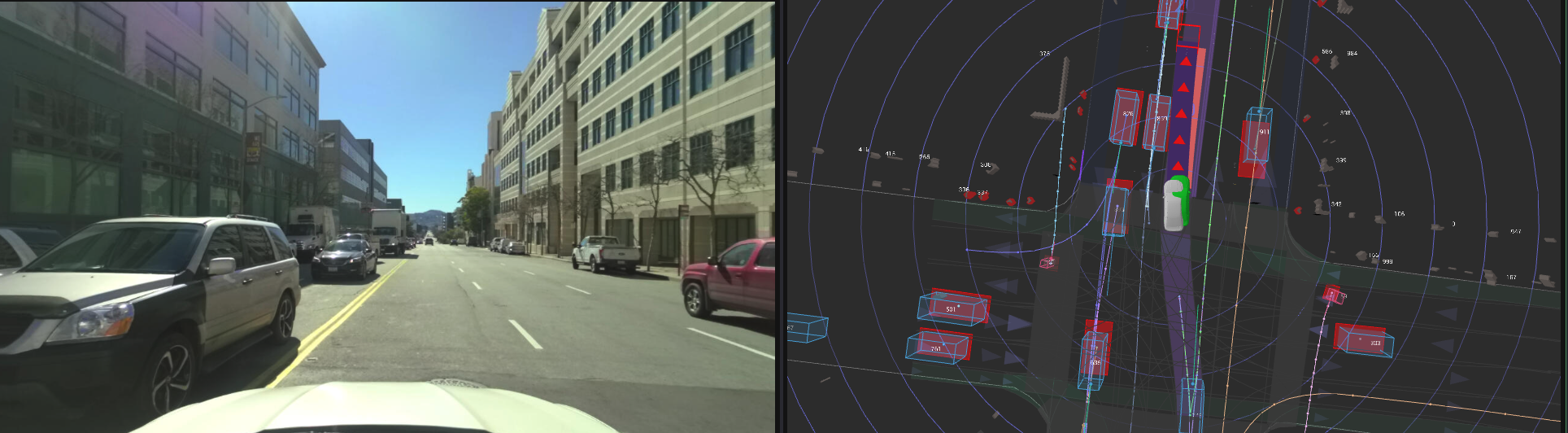}
    \caption{A SafetyNet failure case in simulation. Two simultaneous situations (SDV driving close to dividing line, and a sudden change in velocity of the oncoming vehicle) combine to reveal an issue in the implementation of the system.}
    \label{fig:fallback_failure}
    \vspace{-3mm}
\end{figure}

There is still a small proportion of the ML planner failures that are not caught by SafetyNet. This is mostly because the current fallback layer implementation does not limit the proximity of the SDV to lane boundaries, and does not fully compensate for the uncertainty in the prediction of behavior of other agents. In Fig.~\ref{fig:fallback_failure} we see that the most recent feasible ML trajectory has brought the SDV close to the lane boundary and simultaneously the oncoming vehicle has a sudden change of velocity. With this combination of factors, the fallback trajectory fails to assure a comfortable lateral distance from the oncoming vehicle.

\begin{figure}[t]
    \centering
    \includegraphics[width=\linewidth]{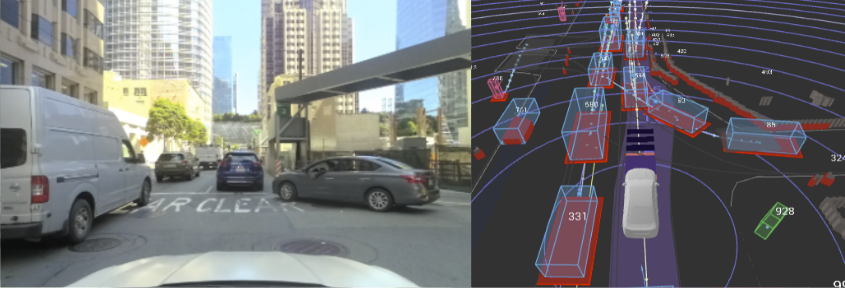}\\
    \vspace{1mm}
    \includegraphics[width=\linewidth]{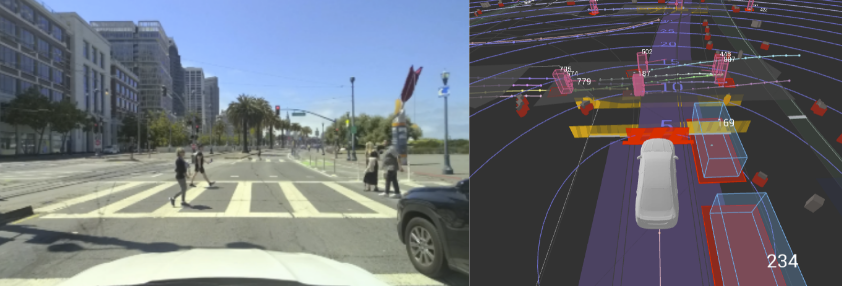}\\
    \vspace{1mm}
    \includegraphics[width=\linewidth]{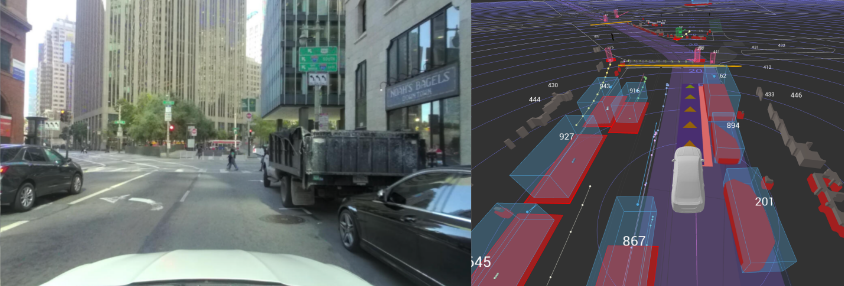}
    \caption{Examples of successful ML planner behavior (no fallback trajectory was used) in the real world. \emph{Top:} merging, \emph{middle:} yielding to pedestrians, and \emph{bottom:} nudging around a parked car.}
    \label{fig:ML_Planner_success_cases}
    \vspace{-3mm}
\end{figure}

\subsection{Real world testing}
Finally, we extensively tested SafetyNet with ML planner (trained on 300 hours), in the real world, in densely populated downtown San Francisco, under the supervision of human safety drivers. During the 150+ mile public road testing, the model successfully performed a wide variety of challenging maneuvers including lane-following, merging, yielding to pedestrians or nudging around parked cars (Fig.~\ref{fig:ML_Planner_success_cases}). At the same time, the fallback layer assured the safety of the overall system, taking control for around 7.9\% of the total driving time. This confirms our hypothesis that although the ML planner is able to perform complex maneuvers and drive safely for most of the time ($>$90\%), an additional layer of safety is required in certain situations. We refer the reader to the accompanying video for a more thorough analysis.

\section{Conclusions}
We present SafetyNet, a method for combining ML planners with a rule-based system fallback layer to provide safe driving in challenging, real-world urban environments. We have demonstrated the very significant improvements in safety and comfort metrics compared to a purely machine-learning-based system, both in simulation as well as in challenging San Francisco streets. This approach makes it possible to safely use learned planners in the real world, and benefit from their ability to improve with the data and handle more complex situations than their purely rule-based counterparts. We believe that teams starting to explore ML planning methods, and even those with state-of-the-art ML planning solutions \cite{bansal2018planning-5, Hawke2020UrbanDW, zeng2019end} would benefit from incorporating a SafetyNet system. In effect, SafetyNet may facilitate the development of machine-learning-based planners and their wider adoption in the AV industry.

We see many exciting opportunities for further development. The fallback layer can be refined to be less conservative and not increase passiveness. In terms of the ML planner, the presented approach is relatively simple, based on imitation learning. It can be improved by drawing from recent advancements in model-based Reinforcement Learning (RL) \cite{moerland2020model}, offline RL \cite{levine2020offline}, or closed-loop training in a data-driven simulation \cite{bergamini2021simnet}.

\section{Acknowledgements}
We would like to thank following contributors who work on Autonomy 2.0 at Level 5:
Luca Bergamini, Sergey Zagoruyko, Ana Ferreira, Oliver Scheel, Stefano Pini, Christian Perone, Lukas Platinsky, Jasper Friedrichs, Jared Wood, Yilun Chen and Alex Ozark.

\addtolength{\textheight}{-3cm}

\bibliographystyle{IEEEtran}
\bibliography{references}

\begin{thebibliography}{10}
\providecommand{\url}[1]{#1}
\csname url@samestyle\endcsname
\providecommand{\newblock}{\relax}
\providecommand{\bibinfo}[2]{#2}
\providecommand{\BIBentrySTDinterwordspacing}{\spaceskip=0pt\relax}
\providecommand{\BIBentryALTinterwordstretchfactor}{4}
\providecommand{\BIBentryALTinterwordspacing}{\spaceskip=\fontdimen2\font plus
\BIBentryALTinterwordstretchfactor\fontdimen3\font minus
  \fontdimen4\font\relax}
\providecommand{\BIBforeignlanguage}[2]{{%
\expandafter\ifx\csname l@#1\endcsname\relax
\typeout{** WARNING: IEEEtran.bst: No hyphenation pattern has been}%
\typeout{** loaded for the language `#1'. Using the pattern for}%
\typeout{** the default language instead.}%
\else
\language=\csname l@#1\endcsname
\fi
#2}}
\providecommand{\BIBdecl}{\relax}
\BIBdecl

\bibitem{jain2021autonomy}
A.~Jain, L.~D. Pero, H.~Grimmett, and P.~Ondruska, ``Autonomy 2.0: Why is
  self-driving always 5 years away?'' 2021.

\bibitem{bansal2018planning-5}
M.~Bansal, A.~Krizhevsky, and A.~Ogale, ``Chauffeurnet: Learning to drive by
  imitating the best and synthesizing the worsts,'' in \emph{Proceedings of
  Robotics: Science and Systems XV}, 2019.

\bibitem{Gao_2020_CVPR}
J.~Gao, C.~Sun, H.~Zhao, Y.~Shen, D.~Anguelov, C.~Li, and C.~Schmid,
  ``Vector{N}et: Encoding {HD} maps and agent dynamics from vectorized
  representation,'' in \emph{IEEE/CVF Conference on Computer Vision and Pattern
  Recognition (CVPR)}, June 2020.

\bibitem{montemerlo2009}
M.~Montemerlo, J.~Becker, S.~Bhat, H.~Dahlkamp, D.~Dolgov, S.~Ettinger,
  D.~Hähnel, T.~Hilden, G.~Hoffmann, B.~Huhnke, D.~Johnston, S.~Klumpp,
  D.~Langer, A.~Levandowski, J.~Levinson, J.~Marcil, D.~Orenstein, J.~Paefgen,
  I.~Penny, and S.~Thrun, ``Junior: The stanford entry in the urban
  challenge.'' 2009, pp. 91--123.

\bibitem{Buehler2009}
M.~Buehler, K.~Iagnemma, and S.~Singh, \emph{The DARPA Urban Challenge:
  Autonomous Vehicles in City Traffic}.\hskip 1em plus 0.5em minus 0.4em\relax
  Springer, 2009.

\bibitem{paden2016survey}
B.~Paden, M.~{\v{C}}{\'a}p, S.~Z. Yong, D.~Yershov, and E.~Frazzoli, ``A survey
  of motion planning and control techniques for self-driving urban vehicles,''
  \emph{IEEE Transactions on intelligent vehicles}, vol.~1, no.~1, pp. 33--55,
  2016.

\bibitem{fan2018}
H.~Fan, F.~Zhu, C.~Liu, L.~Zhang, L.~Zhuang, D.~Li, W.~Zhu, J.~Hu, H.~Li, and
  Q.~Kong, ``Baidu apollo em motion planner,'' 07 2018.

\bibitem{Ajanovic2018}
Z.~Ajanovic, B.~Lacevic, B.~Shyrokau, M.~Stolz, and M.~Horn, ``Search-based
  optimal motion planning for automated driving,'' \emph{2018 IEEE/RSJ
  International Conference on Intelligent Robots and Systems (IROS)}, pp.
  4523--4530, 2018.

\bibitem{KuffnerRRT-connect2000}
J.~J.~K. Jr. and S.~M. Lavalle, ``Rrt-connect: An efficient approach to
  single-query path planning,'' in \emph{Int. Conf. on Robotics and
  Automation}, 2000.

\bibitem{karaman2011}
S.~Karaman and E.~Frazzoli, ``Sampling-based algorithms for optimal motion
  planning,'' \emph{The International Journal of Robotics Research}, vol.~30,
  no.~7, pp. 846--894, 2011.

\bibitem{Bandyopadhyay13}
T.~Bandyopadhyay, K.~S. Won, E.~Frazzoli, D.~Hsu, W.~S. Lee, and D.~Rus,
  ``Intention-aware motion planning,'' in \emph{Algorithmic Foundations of
  Robotics X}, E.~Frazzoli, T.~Lozano-Perez, N.~Roy, and D.~Rus, Eds., 2013.

\bibitem{wei2014behavioral}
J.~Wei, J.~M. Snider, T.~Gu, J.~M. Dolan, and B.~Litkouhi, ``A behavioral
  planning framework for autonomous driving,'' in \emph{2014 IEEE Intelligent
  Vehicles Symposium Proceedings}.\hskip 1em plus 0.5em minus 0.4em\relax IEEE,
  2014, pp. 458--464.

\bibitem{chang2021systems}
Y.~Chang, S.~Omari, and M.~S. Vitelli, ``Systems and methods for determining
  vehicle trajectories directly from data indicative of human-driving
  behavior,'' Jun.~10 2021, uS Patent App. 16/706,307.

\bibitem{bojarski2016end2end}
M.~Bojarski, D.~D. Testa, D.~Dworakowski, B.~Firner, B.~Flepp, P.~Goyal, L.~D.
  Jackel, M.~Monfort, U.~Muller, J.~Zhang, X.~Zhang, J.~Zhao, and K.~Zieba,
  ``End to end learning for self-driving cars,'' 2016.

\bibitem{zeng2019end}
W.~Zeng, W.~Luo, S.~Suo, A.~Sadat, B.~Yang, S.~Casas, and R.~Urtasun,
  ``End-to-end interpretable neural motion planner,'' in \emph{IEEE
  International Conference on Computer Vision and Pattern Recognition (CVPR)},
  2019.

\bibitem{Hawke2020UrbanDW}
J.~Hawke, R.~Shen, C.~Gurau, S.~Sharma, D.~Reda, N.~Nikolov, P.~Mazur,
  S.~Micklethwaite, N.~Griffiths, A.~Shah, and A.~Kendall, ``Urban driving with
  conditional imitation learning,'' \emph{2020 IEEE International Conference on
  Robotics and Automation (ICRA)}, pp. 251--257, 2020.

\bibitem{Pomerleau1989}
D.~A. Pomerleau, ``Alvinn: An autonomous land vehicle in a neural network,'' in
  \emph{Advances in Neural Information Processing Systems}, vol.~1, 1989.

\bibitem{Ross2011ARO}
S.~Ross, G.~J. Gordon, and J.~Bagnell, ``A reduction of imitation learning and
  structured prediction to no-regret online learning,'' in \emph{International
  Conference on Artificial Intelligence and Statistics}, 2011.

\bibitem{shalev2016safe}
S.~Shalev-Shwartz, S.~Shammah, and A.~Shashua, ``Safe, multi-agent,
  reinforcement learning for autonomous driving,'' \emph{arXiv preprint
  arXiv:1610.03295}, 2016.

\bibitem{riedmiller2007learning}
M.~Riedmiller, M.~Montemerlo, and H.~Dahlkamp, ``Learning to drive a real car
  in 20 minutes,'' in \emph{2007 Frontiers in the Convergence of Bioscience and
  Information Technologies}.\hskip 1em plus 0.5em minus 0.4em\relax IEEE, 2007,
  pp. 645--650.

\bibitem{kendall2019learning}
A.~Kendall, J.~Hawke, D.~Janz, P.~Mazur, D.~Reda, J.-M. Allen, V.-D. Lam,
  A.~Bewley, and A.~Shah, ``Learning to drive in a day,'' in
  \emph{International Conference on Robotics and Automation (ICRA)}.\hskip 1em
  plus 0.5em minus 0.4em\relax IEEE, 2019, pp. 8248--8254.

\bibitem{ziebart2008maximum}
B.~D. Ziebart, A.~L. Maas, J.~Bagnell, and A.~Dey, ``Maximum entropy inverse
  reinforcement learning,'' in \emph{AAAI}, 2008.

\bibitem{Wulfmeier2015MaximumED}
M.~Wulfmeier, P.~Ondruska, and I.~Posner, ``Maximum entropy deep inverse
  reinforcement learning,'' \emph{arXiv: Learning}, 2015.

\bibitem{qureshi2019MPNetworks}
A.~H. Qureshi, A.~Simeonov, M.~J. Bency, and M.~C. Yip, ``Motion planning
  networks,'' in \emph{2019 International Conference on Robotics and Automation
  (ICRA)}, 2019, pp. 2118--2124.

\bibitem{arslan2015GuidingRRT}
O.~Arslan and P.~Tsiotras, ``Machine learning guided exploration for
  sampling-based motion planning algorithms,'' in \emph{2015 IEEE/RSJ
  International Conference on Intelligent Robots and Systems (IROS)}, 2015, pp.
  2646--2652.

\bibitem{chaing2019RLRRT}
H.-T.~L. Chiang, J.~Hsu, M.~Fiser, L.~Tapia, and A.~Faust, ``Rl-rrt:
  Kinodynamic motion planning via learning reachability estimators from rl
  policies,'' \emph{IEEE Robotics and Automation Letters}, vol.~4, no.~4, pp.
  4298--4305, 2019.

\bibitem{pulver2020pilot}
H.~Pulver, F.~Eiras, L.~Carozza, M.~Hawasly, S.~Albrecht, and S.~Ramamoorthy,
  ``Pilot: Efficient planning by imitation learning and optimisation for safe
  autonomous driving,'' \emph{arXiv preprint arXiv:2011.00509}, 2020.

\bibitem{casas2021mp3}
S.~Casas, A.~Sadat, and R.~Urtasun, ``Mp3: A unified model to map, perceive,
  predict and plan,'' in \emph{IEEE/CVF International Conference on Computer
  Vision and Pattern Recognition}, 2021, pp. 14\,403--14\,412.

\bibitem{Shalev2017RSS}
\BIBentryALTinterwordspacing
S.~Shalev{-}Shwartz, S.~Shammah, and A.~Shashua, ``On a formal model of safe
  and scalable self-driving cars,'' \emph{CoRR}, vol. abs/1708.06374, 2017.
  [Online]. Available: \url{http://arxiv.org/abs/1708.06374}
\BIBentrySTDinterwordspacing

\bibitem{althoff2014Reachability}
M.~Althoff and J.~M. Dolan, ``Online verification of automated road vehicles
  using reachability analysis,'' \emph{IEEE Transactions on Robotics}, vol.~30,
  no.~4, pp. 903--918, 2014.

\bibitem{qi2016pointnet}
C.~R. Qi, H.~Su, K.~Mo, and L.~J. Guibas, ``Point{N}et: Deep learning on point
  sets for 3d classification and segmentation,'' \emph{arXiv preprint
  arXiv:1612.00593}, 2016.

\bibitem{NIPS2017_3f5ee243}
A.~Vaswani, N.~Shazeer, N.~Parmar, J.~Uszkoreit, L.~Jones, A.~N. Gomez, L.~u.
  Kaiser, and I.~Polosukhin, ``Attention is all you need,'' in \emph{Advances
  in Neural Information Processing Systems}, vol.~30, 2017.

\bibitem{cui2020deep}
H.~Cui, T.~Nguyen, F.-C. Chou, T.-H. Lin, J.~Schneider, D.~Bradley, and
  N.~Djuric, ``Deep kinematic models for kinematically feasible vehicle
  trajectory predictions,'' 2020.

\bibitem{werling2012optimal}
M.~Werling, S.~Kammel, J.~Ziegler, and L.~Gr{\"o}ll, ``Optimal trajectories for
  time-critical street scenarios using discretized terminal manifolds,''
  \emph{The International Journal of Robotics Research}, vol.~31, no.~3, pp.
  346--359, 2012.

\bibitem{dehaan2019causal}
P.~de~Haan, D.~Jayaraman, and S.~Levine, ``Causal confusion in imitation
  learning,'' in \emph{Advances in Neural Information Processing Systems},
  vol.~32, 2019.

\bibitem{moerland2020model}
T.~M. Moerland, J.~Broekens, and C.~M. Jonker, ``Model-based reinforcement
  learning: A survey,'' \emph{arXiv preprint arXiv:2006.16712}, 2020.

\bibitem{levine2020offline}
S.~Levine, A.~Kumar, G.~Tucker, and J.~Fu, ``Offline reinforcement learning:
  Tutorial, review, and perspectives on open problems,'' \emph{arXiv preprint
  arXiv:2005.01643}, 2020.

\bibitem{bergamini2021simnet}
L.~Bergamini, Y.~Ye, O.~Scheel, L.~Chen, C.~Hu, L.~Del~Pero, B.~Osinski,
  H.~Grimmett, and P.~Ondruska, ``Simnet: Learning reactive self-driving
  simulations from real-world observations,'' in \emph{International Conference
  on Robotics and Automation (ICRA)}.\hskip 1em plus 0.5em minus 0.4em\relax
  IEEE, 2021.

\end{thebibliography}

\end{document}